%% file: ZedNet.tex
\documentclass[10pt,twocolumn,letterpaper]{article}

\usepackage{tikz-cd} 
\usepackage{cvpr}
\usepackage{times}
\usepackage{epsfig}
\usepackage{xcolor}
\usepackage{tabularx,booktabs}

\usepackage{calrsfs} 
\usepackage{amsmath}
\usepackage{amssymb}
\usepackage{multirow}
\usepackage[us,12hr]{datetime}
\usepackage{textcomp}
\usepackage{algorithmic}
\usepackage{authblk}
\usepackage{soul}

\usepackage{fancyhdr,graphicx,amsmath,amssymb}
\usepackage[]{algorithm2e}

\DeclareMathAlphabet{\pazocal}{OMS}{zplm}{m}{n}

\usepackage[pagebackref=true,breaklinks=true,letterpaper=true,colorlinks,bookmarks=false]{hyperref}

\cvprfinalcopy 

\def\cvprPaperID{3250} 

\ifcvprfinal\fi

\definecolor{ambrish_color}{RGB}{56,108,176} 
\definecolor{amit_color}{RGB}{217,95,2}    
\definecolor{rohith_color}{RGB}{228,26,28}   
\definecolor{ching_color}{RGB}{80,228,80}   

\newcolumntype{Y}{>{\centering\arraybackslash}X}
\newcolumntype{C}[1]{>{\hsize=#1cm\hsize\centering\arraybackslash}X}%

\makeatletter
\DeclareRobustCommand\onedot{\futurelet\@let@token\@onedot}
\def\@onedot{\ifx\@let@token.\else.\null\fi\xspace}
\def\eg{\emph{e.g}\onedot}

\def\etc{\emph{etc}\onedot} 
 
\def\etal{\emph{et al}\onedot}

\makeatother

\makeatletter
\renewcommand\AB@affilsepx{, \protect\Affilfont}
\makeatother

\begin{document}

\title{Unsupervised 3D Pose Estimation with Geometric Self-Supervision \vspace{-0.3cm}} 

%

\author{Ching-Hang Chen$^1$\qquad Ambrish Tyagi$^1$\qquad Amit Agrawal$^1$\qquad Dylan Drover$^1$\\ \vspace{-10pt}
	Rohith MV$^1$\qquad Stefan Stojanov$^{1,2}$\qquad James M. Rehg$^{1,2}$\qquad
	\vspace{6pt}\\
	$^1$Amazon Lab126, $^2$Georgia Institute of Technology\\
	{\tt\small \{chinghc, ambrisht, aaagrawa, droverd, kurohith\}@amazon.com \{sstojanov, rehg\}@gatech.edu}
	\vspace{-15pt}
}



%
%

\maketitle
\input{abstract}

\input{introduction}

\input{related_work}
\input{algo}

\input{experiments}

\input{conclusions}

\clearpage

\bibliographystyle{ieee_fullname}
\bibliography{references,poseRef,ching}

\end{document}

%% file: abstract.tex
\begin{abstract}
\label{sect:abstract}
We present an unsupervised learning approach to recover 3D human pose from 2D skeletal joints extracted from a single image. Our method does not require any multi-view image data, 3D skeletons, correspondences between 2D-3D points, or use previously learned 3D priors during training. A lifting network accepts 2D landmarks as inputs and generates a corresponding 3D skeleton estimate. During training, the recovered 3D skeleton is reprojected on random camera viewpoints to generate new `synthetic' 2D poses. By lifting the synthetic 2D poses back to 3D and re-projecting them in the original camera view, we can define self-consistency loss both in 3D and in 2D. The training can thus be self supervised by exploiting the geometric self-consistency of the lift-reproject-lift process. We show that self-consistency alone is not sufficient to generate realistic skeletons, however adding a 2D pose discriminator enables the lifter to output valid 3D poses. Additionally, to learn from 2D poses `in the wild', we train an unsupervised 2D domain adapter network to allow for an expansion of 2D data. This improves results and demonstrates the usefulness of 2D pose data for unsupervised 3D lifting. Results on Human3.6M dataset for 3D human pose estimation demonstrate that our approach improves upon the previous unsupervised methods by 30\% and outperforms many weakly supervised approaches that explicitly use 3D data.
\vspace{-1.5ex}
\vspace{-3ex}
\end{abstract}

%% file: introduction.tex
\section{Introduction}
\label{sect:introduction}
\begin{figure*}[htb]
	\centering
	\includegraphics[width=0.99\linewidth, trim={0.5cm 7cm 0.5cm 5.5cm}, clip]{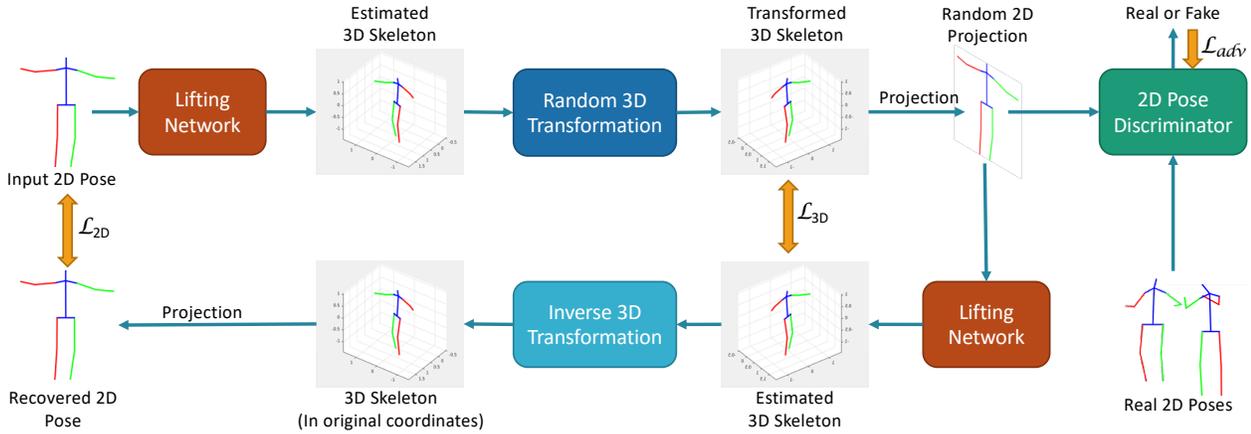}
	\caption{We train a 2D-3D lifting network (lifter), which estimates the 3D skeleton from 2D pose landmarks. Random projections of generated 3D skeletons are fed to a 2D pose discriminator to provide feedback to the lifter. The random projections also go through a similar lifting and reprojection process, allowing the network to self supervise the training process by exploiting geometric consistency.}
	\label{fig:mainarch}
	\vspace{-3ex}
\end{figure*}

Estimation of 3D human pose from images and videos is a classical ill-posed inverse problem in computer vision with numerous applications~\cite{forsyth2006computational,hogg1983model,moeslund2001survey,o1980model} in human tracking, action understanding, human-robot interaction, augmented reality, video gaming,~\etc. Current deep learning-based systems attempt to learn a mapping from RGB images or 2D keypoints to 3D skeleton joints via some form of supervision requiring datasets with \emph{known 3D pose}. However, obtaining 3D motion capture data is time-consuming, difficult, and expensive, and as a result, only a limited amount of 3D data is currently available. On the other hand, 2D image and video data of humans is available in abundance. However, unsupervised learning of 3D joint locations from 2D pose alone remains a holy grail in the field. In this paper, we take a first step towards achieving this goal and present an unsupervised learning algorithm to estimate 3D human pose from 2D pose landmarks/keypoints. Our approach does not use 3D inputs in \textit{any} form and does not require 2D-3D correspondences or explicit 3D priors.

Due to perspective projection ambiguity, there exists an infinite number of 3D skeletons corresponding to a given 2D pose. However, all of these solutions are not physically plausible given the anthropomorphic constraints and joint angle limits of a human body articulation. Typically, supervised learning with 2D pose and corresponding 3D skeletons is used to restrict the solution space. In addition, the 3D structure can also be regularized in a weakly-supervised manner by using priors such as symmetry, ratio of length of various skeleton elements, and kinematic constraints, which are learned from 3D data. In contrast, this paper addresses the fundamental problem of lifting 2D image coordinates to 3D space without the use of any additional cues such as video~\cite{tekin2016direct,Zhou_2016_CVPR}, multi-view cameras~\cite{amin2013multi,hofmann2012multi}, or depth images~\cite{rafi2015semantic,shotton2013real,yub2015random}.

We posit that the following properties of the 2D-3D pose mapping render unsupervised lifting possible: 1) \textit{Closure:} If a 2D skeleton is lifted to 3D accurately, and then randomly rotated and reprojected, the resulting 2D skeleton will lie within the distribution of valid 2D poses. Conversely, a lifted 3D skeleton whose random re-projection falls outside this distribution is likely to be inaccurate. 2) \textit{Invariance:} 2D projections of the same 3D skeleton from different viewpoints, when lifted, should produce the same 3D output. In other words, lifting should be invariant to change in the viewpoint.

We employ the above properties in designing a deep neural network, referred to as the \textit{lifting network}, which is illustrated in Figure~\ref{fig:mainarch}. 
We introduce a novel geometrical consistency loss term that allows the network to learn in a self-supervised mode. This self-consistency loss relies on the property of invariance: any 2D projection of the generated 3D skeleton should produce the same 3D skeleton when processed by the lifting network (Section~\ref{subsect:cycle-loss}). We further demonstrate that self-consistency is a necessary but not a sufficient condition. We add a discriminator to ensure that the projection of lifted skeletons lie within the distribution of 2D poses.  
However we find that self-supervision \textit{does} improve performance of the lifting network when used in conjunction with discriminator feedback.

\textbf{Domain Adaptation:} Since unsupervised learning methods often need more data for training than supervised methods, it is desirable to exploit multiple data sources. However, \textit{domain shifts} could occur in multiple data sources due to (a) differences in human pose and viewpoint variations, and (b) semantic differences in the location of the skeletal joints on the body (e.g., hips marked inside/outside legs).
We propose a domain adaptation algorithm, where we train a 2D \textit{adapter} network to convert 2D joints from a source domain to the target domain without the need for any correspondences.
Using multiple datasets allows us to enrich the viewpoint, pose, and articulation variations in the target domain using additional domains. 

\textbf{Temporal consistency during training:} Our algorithm only requires 2D joints extracted from a single frame for training and inference. However, if sequences of poses from videos are available during training, we show how they can be used to improve the lifter.
To exploit temporal consistency during training, we incorporate an additional \textit{temporal discriminator} that classifies the differences in 2D joints in subsequent frames as real/fake. Our ablation studies show that adding a temporal discriminator improves performance by an additional $7$\%, even when inference is performed on a single frame.

Our paper makes the following contributions:
\vspace{-2ex}
\begin{itemize}
	\setlength{\itemsep}{0pt}
	\setlength{\leftmargin}{0in}
	\item Inspired by ~\cite{ZedNet_2018_ECCVW}, we present an unsupervised algorithm to lift 2D joints to 3D skeletons by observing samples of real 2D poses, without using 3D data in any form.
	\item Our method can learn by exploiting geometric self consistency. We show that self consistency is a necessary but not a sufficient condition for lifting. Self consistency loss improves performance when combined with 2D pose discriminator adversarial loss.
	\item We propose a 2D domain adaptation technique which can utilize data from different domains to improve performance on the target domain.
	\item We show that adding a temporal discriminator during training can further improve performance, even for single frame 2D-3D lifting during inference.
\end{itemize}

%% file: related_work.tex
\section{Related Work}
\label{sect:related_work}

\textbf{3D Pose Estimation:} There are numerous deep learning techniques proposed for estimating 3D joint location directly from 2D images~\cite{orinet,mono-3dhp2017,park20163d,conf/bmvc/ParkK18,Pavlakos_2017_CVPR,Rogez_2017_CVPR}. Other methods decompose this problem into the estimation of 2D joint locations from images followed by the estimation of 3D joint locations based on the 2D keypoints. 2D pose from images can be obtained using techniques such as CPM~\cite{cpm}, Stacked-hourglass architecture~\cite{stacked-hourglass}, Mask-RCNN~\cite{mask-rcnn} or affinity models~\cite{OpenPose}. As discussed, our focus is on estimating 3D pose from 2D landmarks~\cite{ChenDeva2017,Tung_2017_ICCV,MartinezICCV2017}, and we are agnostic to the source of landmarks.
For the purpose of comparison, prior work on lifting can be organized into four categories:
\textbf{Fully Supervised:} These include approaches such as~\cite{Li_2015_ICCV,MartinezICCV2017,Nie_2017_ICCV} that use paired 2D-3D data comprised of ground truth 2D locations of joint landmarks and the corresponding 3D ground truth for learning. For example, Martinez \etal~\cite{MartinezICCV2017} learn a regression network from 2D joints to 3D joints, whereas Moreno-Noguer~\cite{Moreno-Noguer_2017_CVPR} learn a regression from 2D distance matrix to 3D distance matrix using 2D-3D correspondences. Exemplar based methods~\cite{ChenDeva2017,jiang20103d,Yasin_2016_CVPR} use a database/dictionary of 3D skeletons for nearest-neighbor look-up.
Tekin \etal~\cite{Tekin_2017_ICCV} fused 2D and 3D image cues relying on 2D-3D correspondences. Wang~\etal~\cite{DRPose3D} use the 3D ground truth to train an intermediate ranking network to extract the depth ordering of pairwise human joints from a single RGB image. Sun~\etal~\cite{sun2017compositional} use a 3D regression based on bone segments derived from joint locations as opposed to directly using joint locations.
Since these methods model 2D to 3D mappings from a given dataset, they implicitly incorporate dataset-specific parameters such as camera projection matrices, distance of skeleton from the camera, and scale of skeletons. This enables these models to predict metric position of joints {in 3D} on similar datasets, but requires paired 2D-3D correspondences which are difficult to obtain.

\textbf{Weakly Supervised:} Approaches such as~\cite{Brau3DV2016,AAAI18_yxu_3dpose,Tome_2017_CVPR,zhou2017towards,Zhou_2016_CVPR,MonoCap} do not explicitly use paired 2D-3D correspondences, but use \textit{unpaired} 3D data to learn priors on shape (3D basis) or pose (articulation priors). For example, Zhou~\etal~\cite{Zhou_2016_CVPR} use a 3D pose dictionary to learn pose priors and Brau~\etal~\cite{Brau3DV2016} employ an independently trained network that learns a prior distribution over 3D poses (kinematic and self-intersection priors). Tome~\etal~\cite{Tome_2017_CVPR}, Wu~\etal~\cite{InterpreterNetwork2016} and Tung~\etal~\cite{Tung_2017_ICCV} pre-train low-dimensional representations from 3D annotations to obtain priors for plausible 3D poses. Another form of weak supervision is employed by Ronchi~\etal~\cite{relativeposeBMVC18}, where they train a network using relative depth ordering of joints to predict 3D pose from images. Dabral~\etal~\cite{dabral2018learning} uses supervision of 3D skeletons in conjunction with anatomical losses based on joint angle limits and limb symmetry. Rhodin~\etal~\cite{rhodin2018learning} train via 2D data, using multiple images of a single pose in addition to supervision in using 3D data when available. An adversarial training paradigm was used by Yang~\etal~\cite{Yang_2018_CVPR} to improve an existing 3D pose estimation framework, lifting in-the-wild images with no 3D ground truth and comparing them to existing 3D skeletons. 

Similar to our work, the weakly supervised approach of Drover~\etal~\cite{ZedNet_2018_ECCVW} also makes use of 2D projections to learn a 3D prior on human pose. However, Drover~\etal utilize the ground-truth 3D points to generate a large amount (12M) of synthetic 2D joints for training, thus augmenting the original 1.5M 2D poses in Human3.6M by almost $10$ times. This allows them to synthetically over-sample the space of camera variations/angles to learn the 3D priors from those poses. In contrast, we do not use any ground truth 3D projection or 3D data in any form. The fact that we can utilize multiple 2D datasets without any 3D supervision sets us apart from these previous approaches, and enables our method to exploit the large amount of available 2D pose data.



\textbf{Unsupervised:} 
Recently, Rhodin~\etal~\cite{Rhodin_2018_ECCV} proposed an unsupervised method to learn a geometry-aware body representation. Their approach maps one view of the human to another view from a set of given multi-view images. It relies on synchronized multi-view images of subjects to learn an encoding of scene geometry and pose. It also uses video sequences to observe the same subject at multiple time instants to learn appearance. In contrast, we do not require multi-view images or the ability to capture the same pose at multiple time instants. We learn 3D pose from 2D projections alone. Kudo~\etal~\cite{kudo2018unsupervised} present 3D error results (130.9 mm) that are comparable to the trivial baseline reported in ~\cite{ZedNet_2018_ECCVW} (127.3 mm).

\textbf{Learning Using Adversarial Loss:} 
Generative adversarial learning has emerged as a powerful framework for modeling complex data distributions, some use it to learn generative models~\cite{GAN,CyCADA,CycleGANICCV2017}, and~\cite{shat2019rem} leverages it to synthesize hard examples,~\etc. Previous approaches have used adversarial loss for human pose estimation by using a discriminator to differentiate real/fake 2D poses~\cite{Chen_2017_ICCV} and real/fake 3D poses~\cite{Tung_2017_ICCV,Black2018}. To estimate 3D, these techniques still require 3D data or use a prior 3D pose models. In contrast, our approach applies an adversarial loss over randomly projected 2D poses of the generated 3D skeletons. Previous works on image-to-image translation such as CycleGAN~\cite{CycleGANICCV2017} or CyCADA~\cite{CyCADA} also rely on a cycle consistency loss in the image domain to enable unsupervised training. However, we use geometric self-consistency and utilize consistency loss in 3D and 2D joint locations, resulting in a novel method for lifting.

%% file: algo.tex
\section{Unsupervised 2D-3D Lifting}
\label{sect:algo}

In this section, we describe our unsupervised learning approach to lift 2D pose to a 3D skeleton. Let $\textbf{x}_i = (x_i,y_i), i = 1 \ldots N,$ denote $N$ 2D pose landmarks of a skeleton with the root joint (midpoint between hip joints) located at the origin. Let $\textbf{X}_i$ denote the corresponding 3D joint for each 2D joint. We assume a camera with unit focal length centered at the origin $(0,0,0)$. Note that because of the fundamental perspective ambiguity, absolute metric depths cannot be obtained from a single view. Therefore, we fix the distance of the skeleton to the camera to a constant $c$ units. In addition, we normalize the 2D skeletons such that the mean distance from the head joint to the root joint is $\frac{1}{c}$ units in 2D. This ensures that 3D skeleton will be generated with a scale of $\approx1$ unit (head to root joint distance).

\subsection{Lifting Network}
\label{subsect:adversarial-algo}
The lifting network $\textrm{G}(x)$ is a neural network that outputs the 3D joint for each 2D joint.
\begin{equation}
\textrm{G}_{\theta_G}(\textbf{x}) = \textbf{X},
\end{equation}
where $\theta_G$ are the parameters of the lifter learned during training. Internally, the lifter estimates
the depth offset $d_i$ of each joint relative to the fixed plane at $c$ units. The 3D joint is computed as $\textbf{X}_i = (x_iz_i,y_iz_i,z_i)$, where
\begin{equation}
z_i = \max\left(1, c + d_i \right).
\end{equation}

\subsection{Random Projections}
\label{subsect:random_projection-algo}
The generated 3D skeletons are projected to 2D using random camera orientations and these 2D poses are sent to the lifter and discriminator. Let $\textbf{R}$ be a random rotation matrix, created by uniformly sampling an azimuth angle between [-$\pi$, $\pi$] and an elevation angle between [-$\pi$/9, $\pi$/9], and $\textbf{X}_r$ be the location of the root joint of the generated skeleton. The rotated 3D skeleton $\textbf{Y}_i$ is obtained as

\begin{eqnarray}
\textbf{Y}_i = Q(\textbf{X}_i) = \textbf{R} * (\textbf{X}_i - \textbf{X}_r) + \textbf{T},
\end{eqnarray}
where $\textbf{T}=\left[0,0,c\right]$. $Q$ represents the rigid transformation between $\textbf{Y}$ and $\textbf{X}$. The rotated 3D skeleton $\textbf{Y}_i$ is then projected to create a 2D skeleton $\textbf{y}_i=P(\textbf{Y}_i)$, where $P$ denotes perspective projection.

\subsection{Self-Supervision via Loop Closure}
\label{subsect:cycle-loss}
\begin{figure}
\begin{equation*}
\begin{tikzcd}
\textbf{x} \arrow{r}{\textrm{G}(\textbf{x})} &
\textbf{X} \arrow{r}{Q(\textbf{X})} &
\textbf{Y} \arrow{r}{\textrm{P}(\textbf{Y})} &
\textbf{y} \arrow{r}{\textrm{D} (\textbf{y})} \arrow[dl, "\textrm{G}(\textbf{y})" pos=0.75, rounded corners, to path={|- (\tikztotarget) \tikztonodes}] &
\mathtt{real/fake}\\
\textbf{\~x} &
\arrow{l}{\textrm{P}(\textbf{\~X})} \textbf{\~X} &
\arrow{l}{\textrm{Q}^{-1}(\textbf{\~Y})} \textbf{\~Y} &
\end{tikzcd}
\end{equation*}
\caption{Self-supervision achieved by closing the loop between the generated skeleton $\textbf{Y}$, its random projection $\textbf{y}$. The recovered 3D skeleton $\textbf{\~Y}$ is obtained by lifting $\textbf{y}$. Upon reversing the geometric transformations, training can be self-supervised by comparing $\textbf{x}$ with $\textbf{\~x}$, and \textbf{Y} with \textbf{\~Y}.}
\label{fig:cycle-loss-fig}
\end{figure}
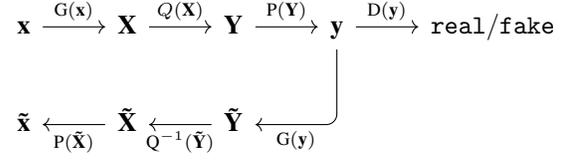

We now describe the symmetrical lifting and projection step performed on the synthesized 2D pose, $\textbf{y}_i$. As shown in Figure~\ref{fig:cycle-loss-fig}, we lift the randomly projected pose $\textbf{y}_i$ to obtain $\textbf{\~Y}_i$
\begin{equation}
\textbf{\~Y}_i = \textrm{G}_{\theta_G}(\textbf{y}_i).
\end{equation}
$\textbf{\~Y}_i$ is transformed to $\textbf{\~X}_i$ by applying the inverse of rigid transformation $Q$ that was used while generating the random projection $\textbf{y}_i$ from $\textbf{X}_i$. The 3D skeleton $\textbf{\~X}_i$ is finally projected to the 2D skeleton $\textbf{\~x}_i$.

Note that the lifting network ${G}(\cdot)$ remains the same in both the forward and backward part of the cycle as illustrated in Figure~\ref{fig:cycle-loss-fig}. If the lifting network accurately reconstructs the 3D pose from 2D inputs, then the 3D skeletons $\textbf{Y}_i$ and $\textbf{\~Y}_i$ and the corresponding 2D projections $\textbf{x}_i$ and $\textbf{\~x}_i$ should be similar. The cycle described herein provides a strong signal for self-supervision for the lifting network, whose loss term can be updated by adding two additional components, namely, $\pazocal{L}_{3D} = \left\lVert{\textbf{Y} - \textbf{\~Y}}\right\rVert^2$ and $\pazocal{L}_{2D} = \left\lVert{\textbf{x} - \textbf{\~x}}\right\rVert^2$.

\subsection{Discriminator for 2D Poses}
\label{subsect:discriminator}
The 2D pose discriminator $D$ is a neural network (with parameters $\theta_D$) that takes as input a 2D pose and outputs a probability between $0$ and $1$. It classifies between real 2D pose $\textbf{r}$ (target probability of $1$) and fake (projected) 2D pose $\textbf{y}$ (target probability of $0$).
Note that for any training sample $\textbf{x}$ for lifter, we do {not} require $\textbf{r}$ to be same as $\textbf{x}$ or any of it's multi-view correspondences. During learning we utilize a standard GAN loss~\cite{GAN} defined as
\begin{equation}
\min_{\theta_G} \max_{\theta_D} \pazocal{L}_{adv} = \mathbb{E}(\log(D(\textbf{r}))) + \mathbb{E}(\log(1-D(\textbf{y}))).
\end{equation}
The discriminator provides feedback to the lifter allowing it to learn priors on 3D skeletons such as the ratio of limb lengths and joint angles using only random 2D projections, thus allowing it to avoid inadequacies as shown in Sect.~\ref{subsect:degeneracy_self_consistency}.

\subsection{Temporal Consistency}
\label{subsect:temporal_consistency}
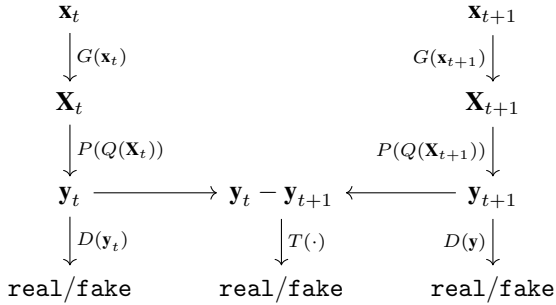
\begin{figure}
\begin{equation*}
\begin{tikzcd}
\textbf{x}_{t} \arrow[d, "G(\textbf{x}_{t})"]   & &\textbf{x}_{t+1} \arrow[d, "G(\textbf{x}_{t+1})", swap] \\
\textbf{X}_{t} \arrow[d, "P(Q(\textbf{X}_{t}))"]  &  & \textbf{X}_{t+1} \arrow[d, "P(Q(\textbf{X}_{t+1}))", swap] \\
\textbf{y}_{t} \arrow[d, "D(\textbf{y}_{t})"] \arrow[r] &  \textbf{y}_{t} - \textbf{y}_{t+1} \arrow[d, "T(\cdot)"] &  \arrow[l]\textbf{y}_{t+1} \arrow[d, "D(\textbf{y})", swap]\\
\mathtt{real/fake} &  \mathtt{real/fake} & \mathtt{real/fake}
\end{tikzcd}
\end{equation*}
\caption{
	Discriminator $T(\cdot)$ enforces a distribution on the temporal differences of projected 2D poses. The temporal consistency is an optional element to stabilize the results, and is only added during training, allowing inference on single frame 2D pose inputs. Subscripts $t$ and $t+1$ denote two consecutive inputs. Lifting, transformation and projection is done as in Figure~\ref{fig:cycle-loss-fig}.}
\label{fig:temporal_consistency}
\end{figure}

Note that our approach does not require video data for training. However, when available, temporal 2D pose sequences (\eg video sequence of actions) can improve the accuracy of the single frame lifting network. We exploit the temporal smoothness via an additional loss function to refine the lifting network $G(\cdot)$ as shown in Figure~\ref{fig:temporal_consistency}. We train an additional discriminator, $T(\cdot)$ that takes as input the difference of 2D poses adjacent in time. The real data for this discriminator comes from a sequence of real 2D poses available during training, $\textbf{r}_t - \textbf{r}_{t+1}$. The discriminator $T(\cdot)$ is updated to optimize the loss that can distinguish the distribution of real 2D pose differences from those of the fake 2D (sequential) projections  $\textbf{y}_t - \textbf{y}_{t+1}$. Specifically,
\begin{equation}
\begin{split}
\max_{\theta_T} \pazocal{L}_T = & \mathbb{E} (\log(T\left(\textbf{r}_t - \textbf{r}_{t+1})\right)) + \\
          & \mathbb{E} (\log(1 - T\left(\textbf{y}_t - \textbf{y}_{t+1})\right)).
\end{split}
\end{equation}

\subsection{Learning from 2D Poses in the Wild}
\label{subsect:domain_adaptation}
\begin{figure}[htb]
	\centering
	\includegraphics[width=0.97\linewidth]{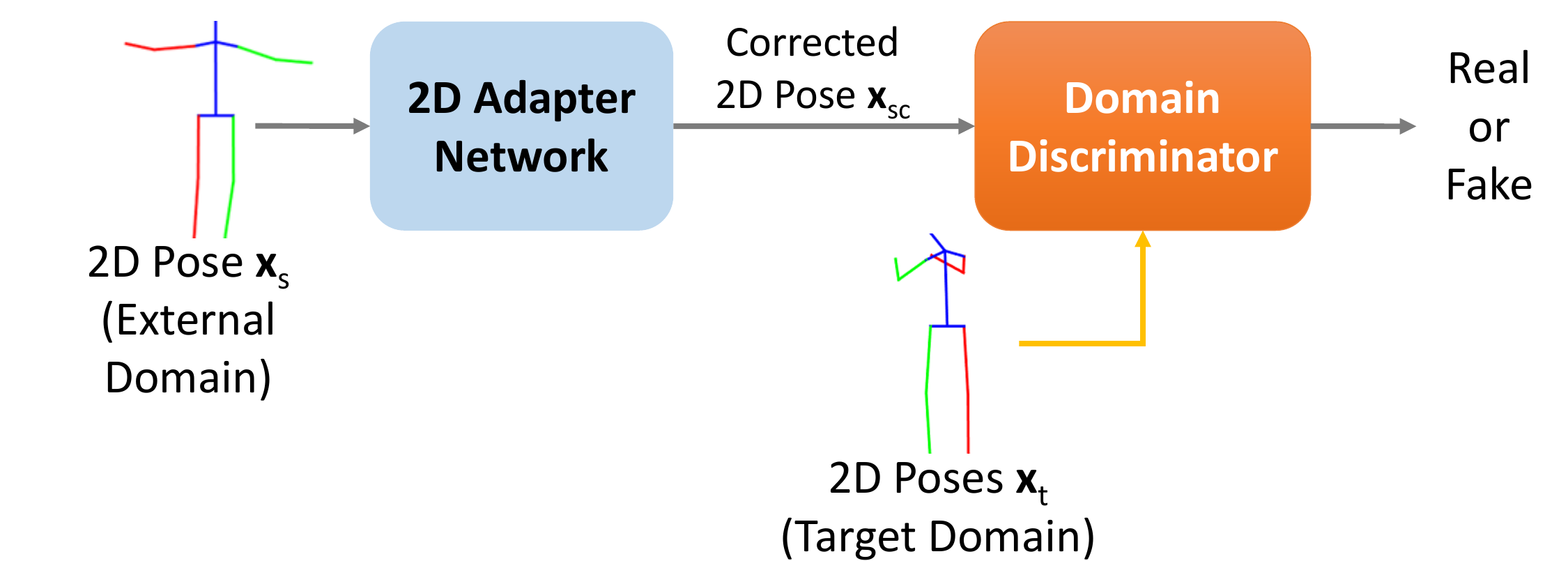}
	\caption{Unsupervised domain adaptation to transform 2D poses from source domain to match the semantics of the target domain. The 2D adapter modifies the input $\textbf{x}_s$ to generate corrected pose $\textbf{x}_{sc}$. The domain discriminator is used to ensure that the distribution of adapted2D poses $\textbf{x}_{sc}$ and the target domain 2D poses $\textbf{x}_t$ match. Adapted poses $\textbf{x}_{sc}$ are used for training a domain agnostic lifter as shown in Figure~\ref{fig:mainarch}.}
	\label{fig:da}
\end{figure}

To improve the 3D lifting accuracy in the target domain of interest (\eg Human 3.6M, $\textbf{x}_t$), we wish to augment 2D training data from in the wild (\eg OpenPose joint estimates on Kinetics dataset, $\textbf{x}_s)$. Depending on the choice of 2D pose extraction algorithms~\cite{OpenPose,stacked-hourglass,cpm}, the position and semantics of 2D keypoints can vary greatly from the representation adopted by the target domain (\eg center of face vs. top of the head or side of the hips vs. pelvis).
	
We train a 2D domain adapter neural network $C$ to map the source domain 2D joints to target domain 2D joints (see Figure~\ref{fig:da}). Let $\textbf{x}_{sc}$ denote the corrected source domain 2D joints, such that $\textbf{x}_{sc} = \textbf{x}_s + C(\textbf{x}_s)$. Note that we do not assume any correspondences between the 2D joints in the source and target domains. Thus, we cannot train $C$ using any form of supervised loss. In absence of any supervision, we use a domain discriminator ${D}_{D}$ to match the distribution between the two domains. Again utilizing the standard GAN loss~\cite{GAN}, we optimize the following loss
\begin{eqnarray}\label{eqn:domain_loss}
\begin{aligned}[b]
\min_{\theta_{C}} \max_{\theta_{D_D}} \pazocal{L}_{adv}  = & \mathbb{E}(\log(D_D(\textbf{x}_p)))  + \lambda \left\|C(\textbf{x}_s)\right\|^2 \\
+ & \mathbb{E}(\log(1-D_D(\textbf{x}_{sc}))),
\end{aligned}
\end{eqnarray}

\noindent where, the $\lambda \left\|C(\textbf{x}_s)\right\|^2$ is a regularizer term to keep the corrections limited to a small magnitude.

\begin{figure}[htb]
	\centering
	\includegraphics[width=0.75\linewidth]{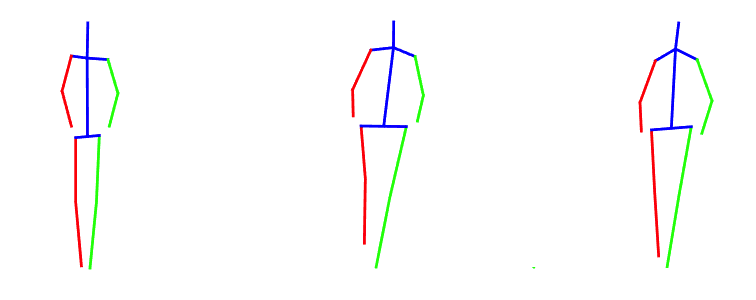}
	\caption{An example of unsupervised domain adaptation. (Left) 2D joints estimated using OpenPose for an example DeepMind Kinetics image. (Middle) Resulting 2D pose after adaptation. (Right) Similar pose from Human3.6M dataset. Notice the change in the width of hips and slant of shoulders after adaptation.}
	\label{fig:domain_corrected_examples}
	\vspace{-3ex}
\end{figure}

Figure~\ref{fig:domain_corrected_examples} shows an example of the difference in semantics between the Human3.6M (target domain) and OpenPose (source domain). In OpenPose, the top of the head is not marked and the center of the marked eye joints are used. In addition, the shoulder keypoints are marked higher than in Human3.6M. The domain adapted 2D pose (middle) is closer in terms of keypoint locations to the target domain. Our domain correction is an off-line preprocessing step. The domain corrected 2D poses, $\textbf{x}_{sc}$, are fed both to the lifting network (Sect.~\ref{subsect:adversarial-algo}) and the 2D pose discriminator (Sect.~\ref{subsect:discriminator}) during training.

\subsection{Training}
As discussed, the 2D to 3D lifting network is trained using geometric self-supervision along with 2D pose and temporal discriminators. Network parameters are updated to optimize the total loss given by,
\begin{equation}\label{eqn:final_loss}
\pazocal{L} = \pazocal{L}_{adv} + w_{2D} \pazocal{L}_{2D} + w_{3D} \pazocal{L}_{3D} + w_T \pazocal{L}_T,
\end{equation}
\noindent where, $w_{2D}=10, w_{3D}=0.001$, and $w_{T}=1$ are the relative weights for each of the 2D, 3D, and temporal loss terms, respectively.


\textbf{Architectures:} We do not use any convolutional layers and use fully connected layers (followed by residual blocks) for all the neural networks described above. Both the lifting network and the 2D pose discriminator takes as input $2N$ dimensional vectors, where $N$ denotes the number of 2D/3D pose points. Similarly, the temporal discriminator takes $2N + 2NM$ inputs corresponding to the pose joints in the current frame and their temporal differences with $M$ other consecutive frames (before and/or after). We adopt a similar architecture as that of Martinez~\etal~\cite{MartinezICCV2017}, with the lifting network composed of 4 residual blocks and the discriminator with 3 residual blocks. For 2D domain adaptation, we use 4 residual blocks for the adapter and 3 residual blocks for the domain discriminator. Batch normalization was used in the lifter and the adapter but not in either of the discriminators.
Our experiments use $N=14$ joint locations. For training we used a batch size of $8192$, a constant depth $c=10$ and the Adam optimizer. 

%% file: experiments.tex
\section{Experimental Evaluation}
\label{sect:experiments}
We present quantitative and qualitative results on the widely used Human3.6M dataset~\cite{h36m} for benchmarking. Additionally, to demonstrate how the unsupervised learning framework can be improved by leveraging 2D pose data from in-the-wild images, we augment our training data by adapting OpenPose estimated 2D human poses from the Kinetics~\cite{kinetics} 
dataset. We also show qualitative visualizations of reconstructed 3D skeletons from 2D pose landmarks on the MPII~\cite{andriluka14cvpr} and Leeds Sports Pose (LSP)~\cite{johnson2010clustered} datasets, for which the ground truth 3D data is not available.


\subsection{Dataset and Metrics}
\label{sect:dataset_metrics}


\noindent\textbf{Human3.6M Dataset:} Human3.6M is one of the largest 3D human pose datasets, consisting of $3.6$ million 3D human poses. The dataset contains video and motion capture (MoCap) data from $5$ female and $6$ male subjects. Data is captured from $4$ different viewpoints, while subjects perform typical activities such as talking on phone, walking, eating,~\etc.

\noindent\textbf{MPI-INF-3DHP:} The MPI-INF-3DHP ~\cite{mono-3dhp2017} is a large human pose dataset containing $>$1.3M frames taken from diverse viewpoints. The dataset has 4 male and 4 female actors performing an array of actions similar to but more diverse than the Human3.6M dataset.

\noindent\textbf{Kinetics dataset:}
The Kinetics dataset contains 400 video clips each for 400 activities involving one or more persons. 
The video clips are sourced from Youtube and each clip is approximately 10 seconds in duration. We did not use any of the class annotations from the dataset for our training. Instead, we extracted 2D pose landmarks using OpenPose~\cite{OpenPose} on sampled frames from this dataset. We retained only those frames in which all the landmarks on a person were estimated with sufficient confidence. 
After this filtering, approximately 9 million 2D skeletons were obtained.

\noindent\textbf{Evaluation Metric:} We report the Mean Per Joint Position Error (MPJPE) in millimeters after scaling and rigid alignment to the ground truth skeleton. Similar to previous works~\cite{ZedNet_2018_ECCVW,Tung_2017_ICCV,li20143d,MartinezICCV2017,Rhodin_2018_ECCV,tekin2016direct,Zhou_2016_CVPR}, we report results on subjects S9 and S11. Also, following the convention as in~\cite{MartinezICCV2017,Rhodin_2018_ECCV}, we only use data from subjects S1, S5, S6, S7, and S8 for training. We do not train class specific models or leverage any motion information during inference to improve the results. The reported metrics are taken from the respective papers for comparisons. 
We also compare our method to ~\cite{mono-3dhp2017, zhou2017towards} which uses the adapted Percentage of Correct Keypoints (PCK) and corresponding Area Under Curve (AUC) metrics.

\input{ResultTable_Abalation}

\subsection{Quantitative Results}
\label{subsect:quant_results}

We summarize our results for Human3.6M and MPI-INF-3DHP in Table~\ref{table:result_summary} and Table~\ref{table:MPI}, respectively. In addition to comparing with the state-of-the-art unsupervised 3D pose estimation method of Rhodin~\etal~\cite{Rhodin_2018_ECCV}, we also show results from top fully supervised and weakly supervised methods. 
Results from~\cite{Rhodin_2018_ECCV} uses images as input and are hence comparable to Ours(SH) results which use 2D joints extracted from the same input images using SH detector~\cite{stacked-hourglass}. Our method reduces error by 30\% compared to~\cite{Rhodin_2018_ECCV} (68mm vs. 98.2mm).

Table~\ref{table:result_ablation} shows the results of an ablation study on lifter with various algorithmic components using ground truth 2D points.~\textbf{SS} denotes self-consistency (Sect.~\ref{subsect:cycle-loss}),  \textbf{Adv} adds the 2D pose discriminator (Sect.~\ref{subsect:discriminator}), \textbf{DA} augments the training data by adapting 2D poses from Kinetics (Sect.~\ref{subsect:domain_adaptation}), and \textbf{TD} leverages temporal cues during training (Sect.~\ref{subsect:temporal_consistency}), when available. As further analyzed in Sect.~\ref{subsect:degeneracy_self_consistency}, just using self consistency loss can lead to unrealistic skeletons without the additional discriminator. Augmenting our approach with additional 2D poses obtained from the Kinetics dataset (Ours: Adv + SS + DA) further reduces the error down to 55mm. Lastly, we exploit temporal information during training (Ours: Adv + SS + DA + TD), when available, to obtain an error of 51mm on Human3.6M. It should be noted that the inference for the TD experiment is still done on single frames and the results can be further improved by applying temporal smoothness techniques on video sequences.
\begin{figure*}[htb]
	\centering
	\includegraphics[width=0.33\linewidth]{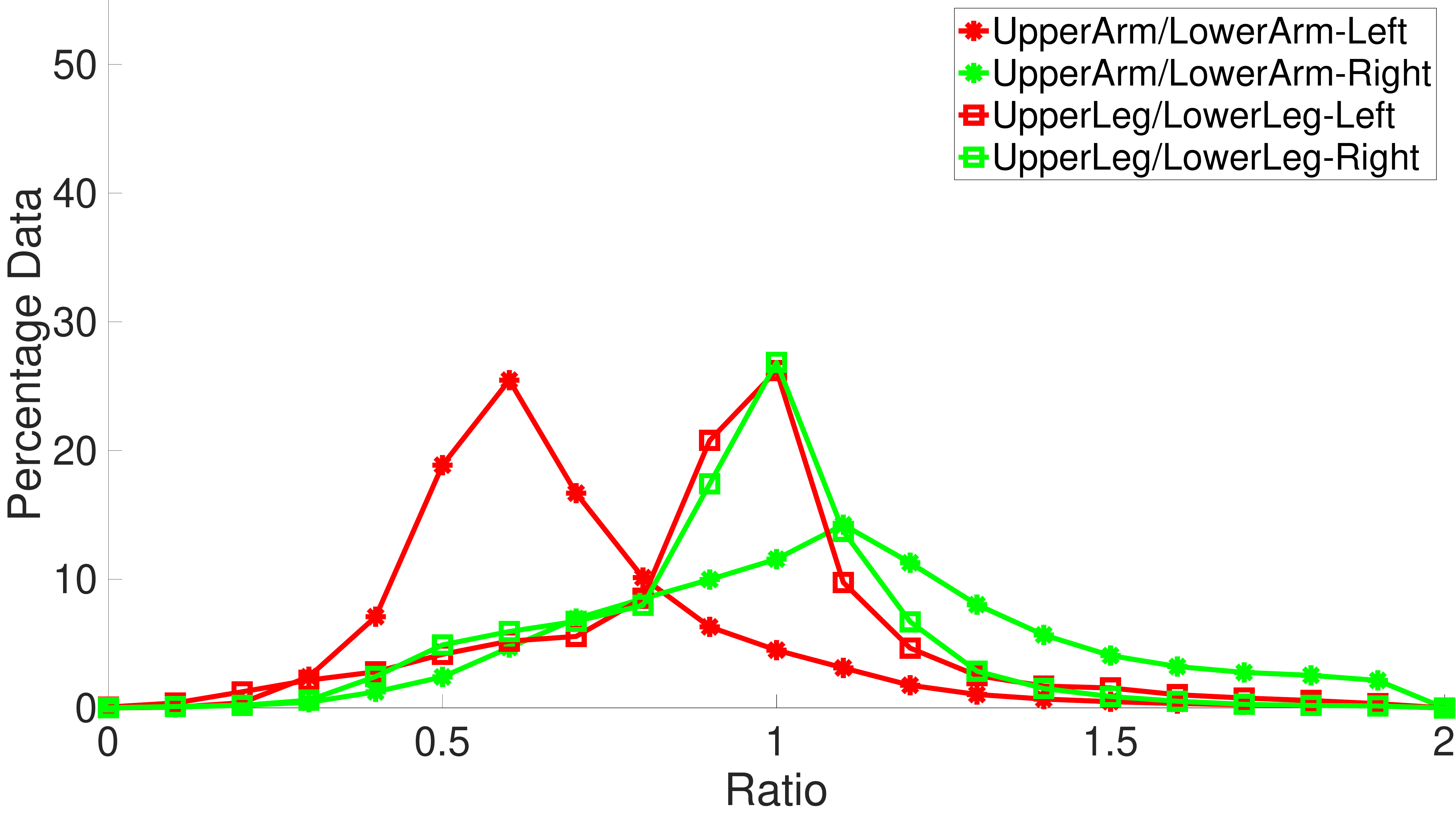}
	\includegraphics[width=0.33\linewidth]{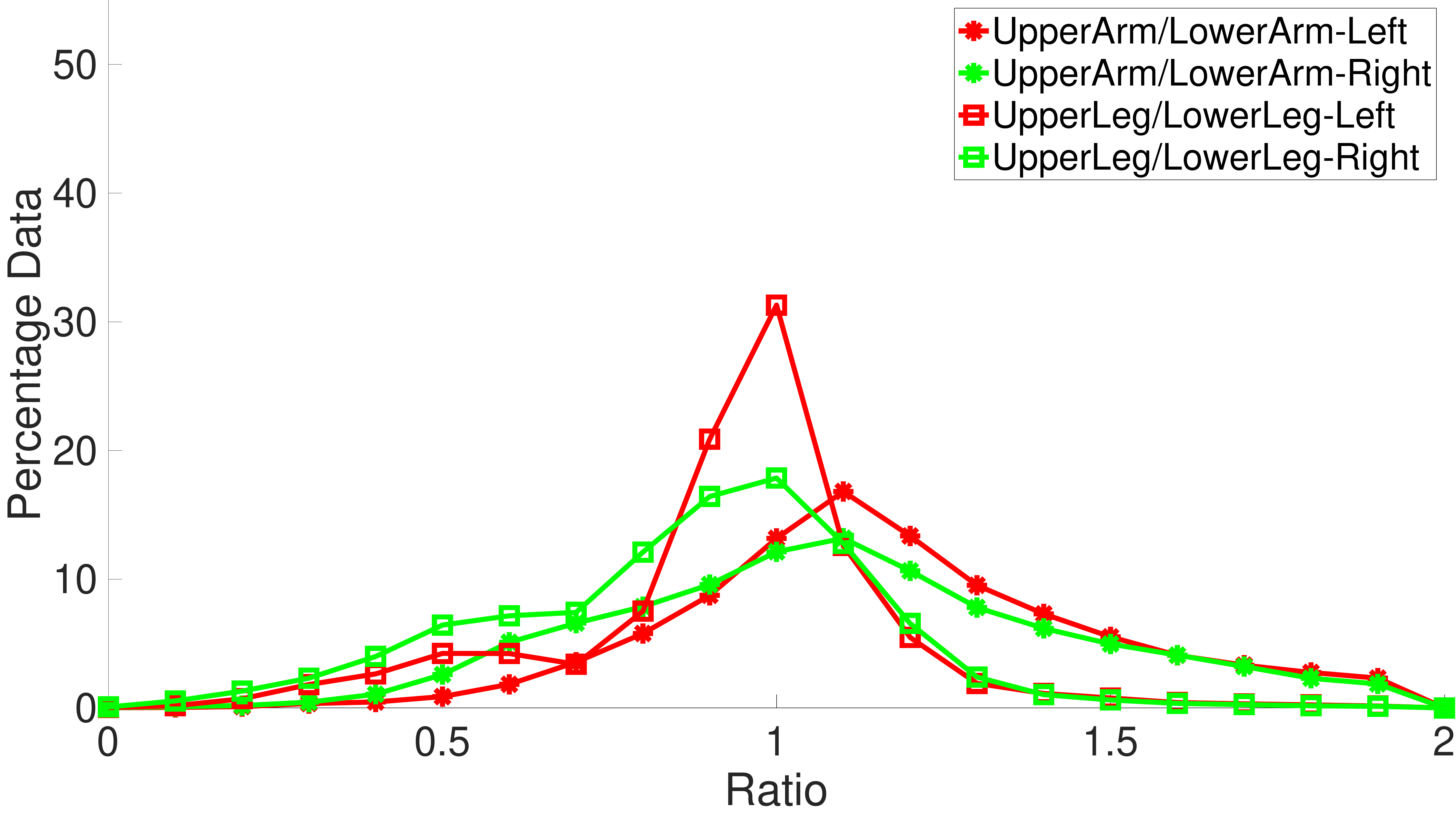}
	\includegraphics[width=0.33\linewidth]{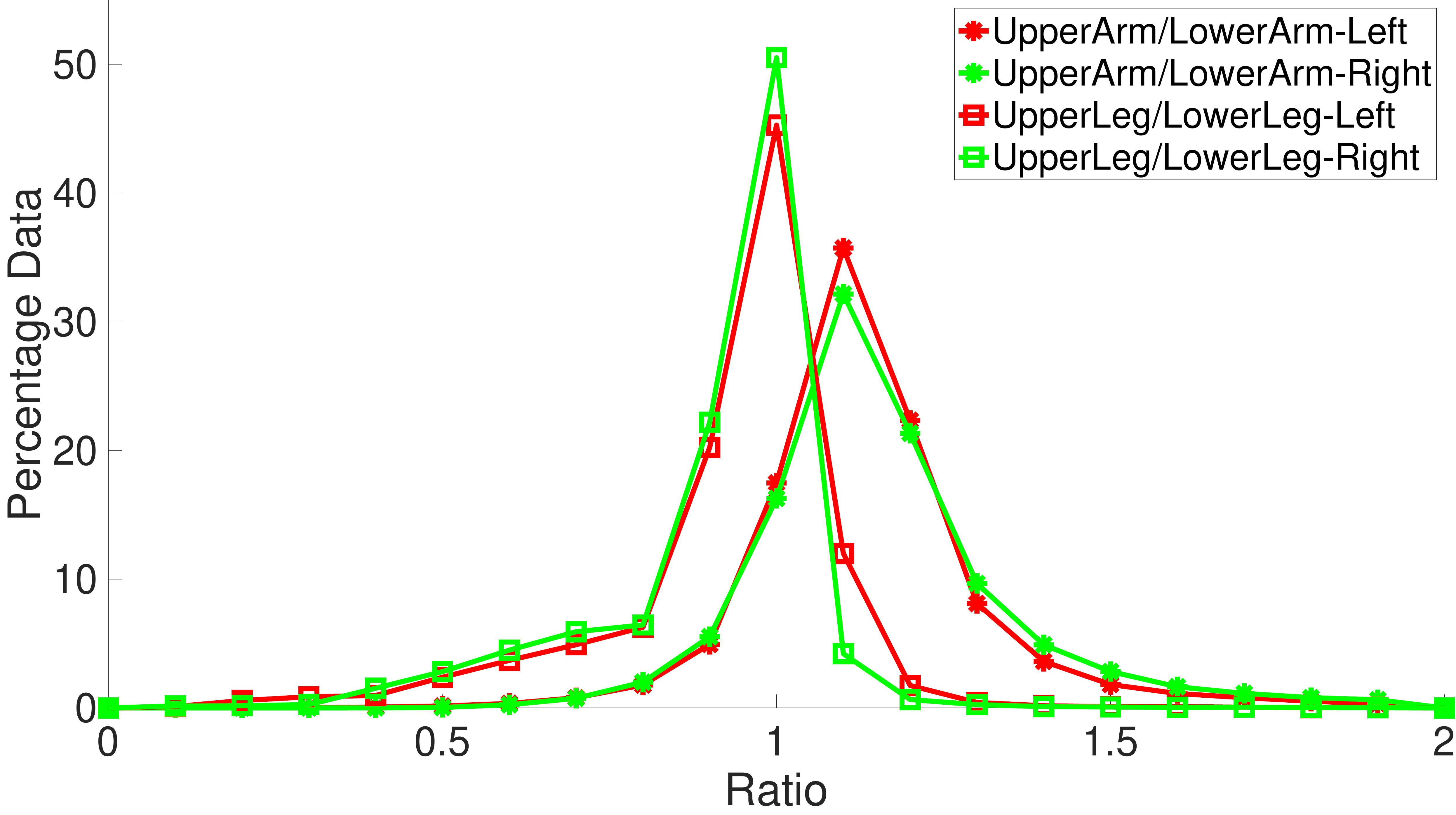}
	\caption{Distribution of limb lengths ratios on Human3.6M test data. (Left) Training using self-consistency loss alone does not impose symmetry. (Middle) Using self-consistency and symmetry aligns distribution of left/right limbs, but results in flatter (unrealistic) distributions. (Right) Using a discriminator sharpens the distributions and brings them closer to real values (ground truth ratio is $\thicksim1.0$ and $\thicksim1.1$ for leg and arms respectively.) }
	\label{fig:selfAnalysis}
\vspace{-4ex}
\end{figure*}
\begin{figure}[htb!]
\centering
\includegraphics[width=0.97\linewidth]{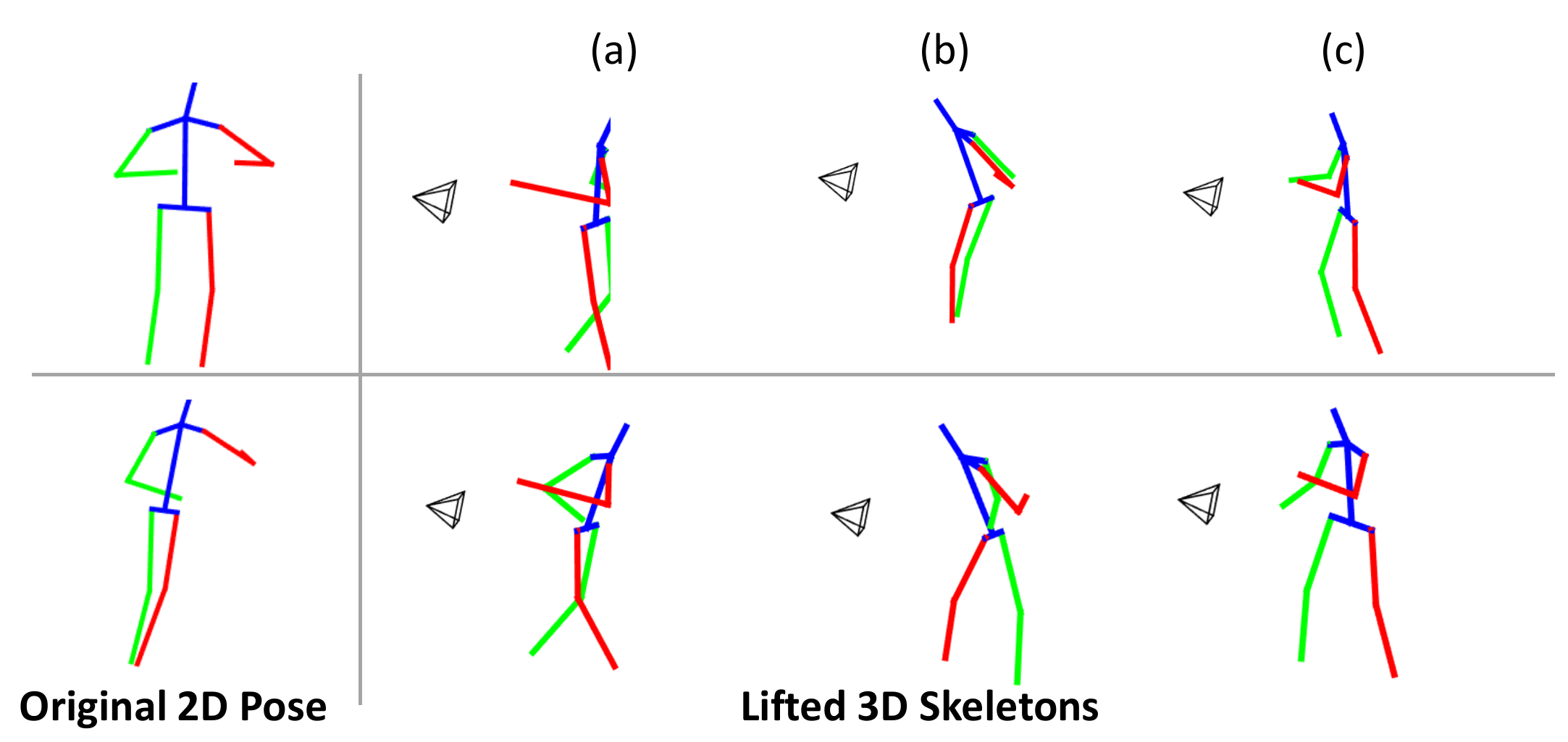}
\caption{Inadequacy of self-consistency loss. Left most Col is input 2D pose. (a) Self-consistency alone is unable to recover the correct 3D skeletons. (b) With symmetry constraints, limb lengths become symmetric but may not have realistic ratios. (c) Adding 2D pose discriminator results in a geometrically consistent and realistic 3D skeletons.}
\label{fig:self_consistency_example}
\end{figure}

\subsection{Inadequacy of Geometric Self-Supervision}
\label{subsect:degeneracy_self_consistency}
At first glance, it may appear that self supervision is sufficient to learn a good lifter, without the need for a discriminator. However, we found that in absence of the 2D pose discriminator, network can produce outputs which are geometrically self-consistent, but not realistic (see Figure~\ref{fig:self_consistency_example}). We present an analysis of the 3D outputs that the lifting network can generate with only self-supervision. Specifically, we examine the ratios of upper to lower arm and leg, both for the left and right side of human body ($4$ ratios).

Figure~\ref{fig:selfAnalysis} (Left) shows the distribution of the $4$ ratios, for a lifter trained using self-consistency loss alone. Note that the lifter produces different limb length ratios for the left and right side of the body. Thus self-consistency loss alone may not produce symmetric (realistic) skeletons without any 3D priors. 
Figure~\ref{fig:selfAnalysis} (Middle) shows that after imposing symmetry constraints, the distributions of the left and right limbs are better aligned. However, the distributions are \textit{flatter} since enforcing the \textit{same} ratios for left and right sides does not ensure that these ratios are \textit{realistic} (conforming to a human body). In other words, the lifter may choose different ratios for different training examples. Figure~\ref{fig:selfAnalysis} (Right) shows the distributions when a discriminator that gives feedback to the lifter using real 2D poses is used. Notice that the ratio distributions become sharper and closer to distributions of real ratios in the training set.  This is the reason that using self-supervision loss (SS) alone performs worse in our ablation studies as shown in Table~\ref{table:result_summary}. However, the self-consistency further improves the performance in conjunction with 2D pose discriminator (Adv+SS).

Note that we do not use symmetry ratios in our framework when the discriminator is present. Our lifting network can learn higher order 3D skeleton statistics (beyond symmetry) based on the feedback from geometric self consistency and the 2D pose discriminator.

\begin{figure}[h!]
	\centering
	\includegraphics[width=0.32\linewidth,trim={2.5cm .7cm 0 .7cm},clip]{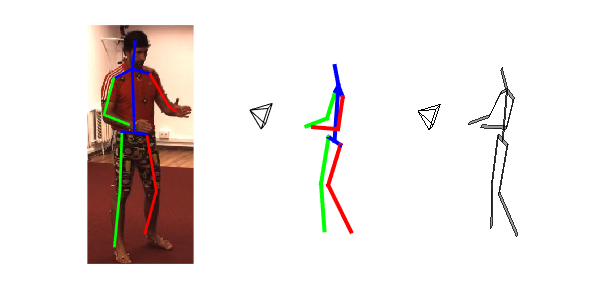}
	\includegraphics[width=0.32\linewidth,trim={2.5cm .7cm 0 .7cm},clip]{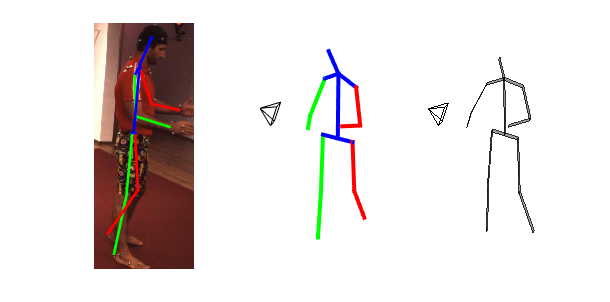}
	\includegraphics[width=0.32\linewidth,trim={2.5cm .7cm 0 .7cm},clip]{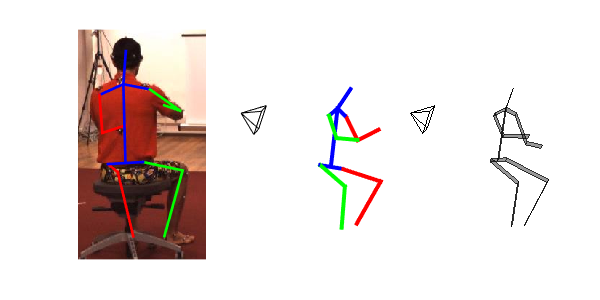}\\
	\includegraphics[width=0.32\linewidth,trim={2.5cm .7cm 0 .7cm},clip]{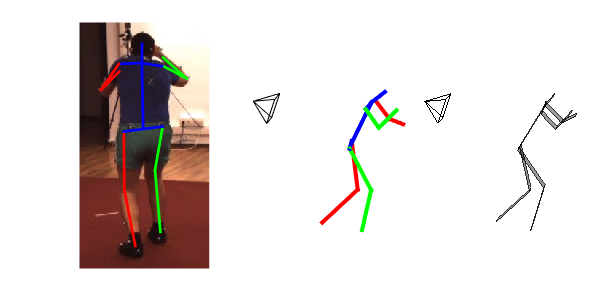}
	\includegraphics[width=0.32\linewidth,trim={2.5cm .7cm 0 .7cm},clip]{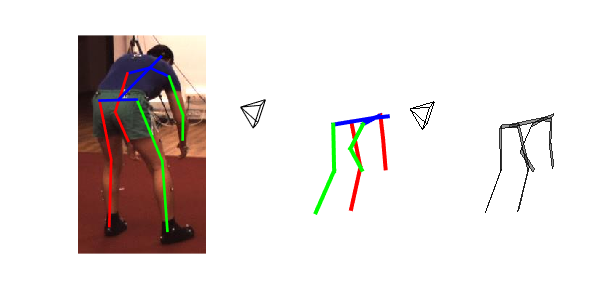}
	\includegraphics[width=0.32\linewidth,trim={2.5cm .7cm 0 .7cm},clip]{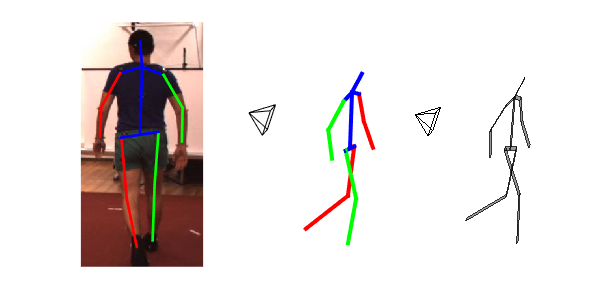}
	\caption{Qualitative results on Human3.6M dataset. (Left to right) Color image with overlaid 2D pose points, estimated and ground-truth 3D skeleton.}
	\label{fig:h36good}
\end{figure}

\begin{figure}[h!]
	\centering
	\includegraphics[width=0.32\linewidth,trim={2.5cm .7cm 0 .7cm},clip]{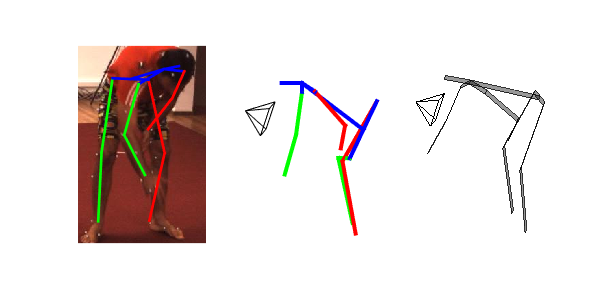}
	\includegraphics[width=0.32\linewidth,trim={2.5cm .7cm 0 .7cm},clip]{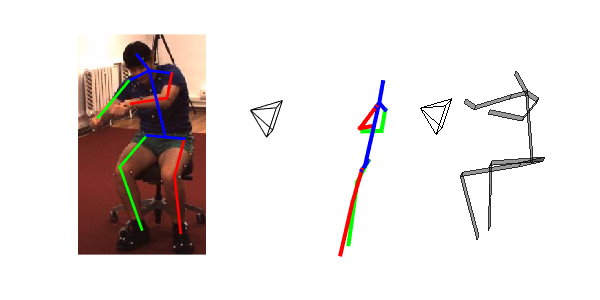}
	\includegraphics[width=0.32\linewidth,trim={2.5cm .7cm 0 .7cm},clip]{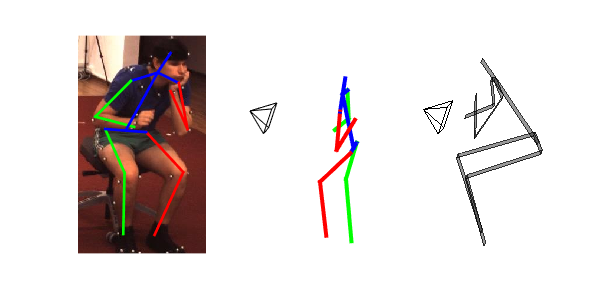}\\
	\includegraphics[width=0.32\linewidth,trim={2.5cm .7cm 0 .7cm},clip]{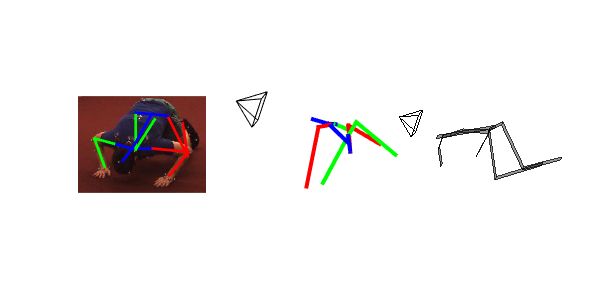}
	\includegraphics[width=0.32\linewidth,trim={2.5cm .7cm 0 .7cm},clip]{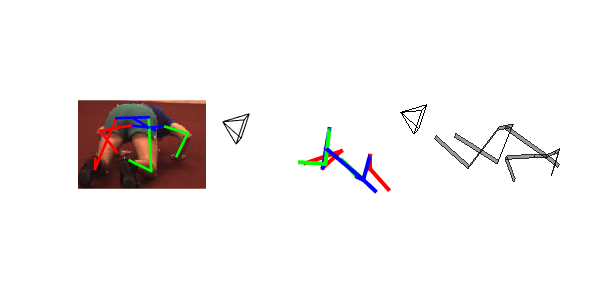}
	\includegraphics[width=0.32\linewidth,trim={2.5cm .7cm 0 .7cm},clip]{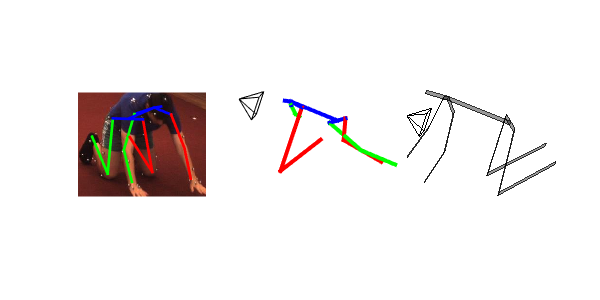}
	\caption{Examples of failure cases of our algorithm on Human3.6M dataset. Ground truth 3D skeletons are shown in gray.}
	\label{fig:h3failure}
\vspace{-3ex}
\end{figure}

\subsection{Semi-supervised 3D Pose Estimation}
Other methods have shown improvement in accuracy when a small amount of 3D data is used for supervised fine-tuning. We fine tuned our baseline model (from unsupervised training) using 5\% of randomly sampled 3D data available in Human3.6M dataset. With this, our method could achieve performance comparable to fully supervised method (37mm) as shown in Table~\ref{table:result_ablation}.

\begin{figure}[htb!]
	\centering
	
	
	
	\includegraphics[width=0.32\linewidth,trim={2.5cm .7cm 0 .7cm},clip]{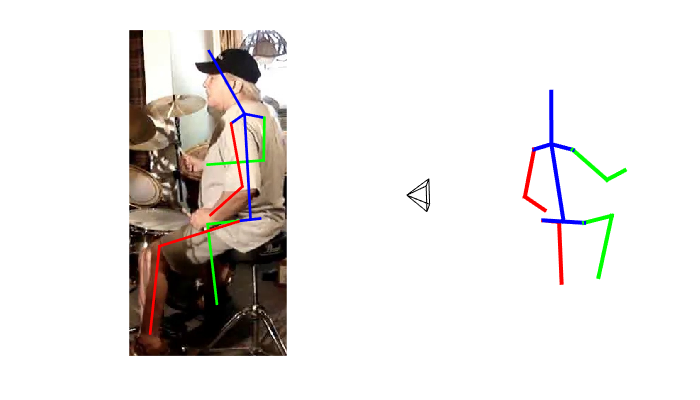}
	\includegraphics[width=0.32\linewidth,trim={2.5cm .7cm 0 .7cm},clip]{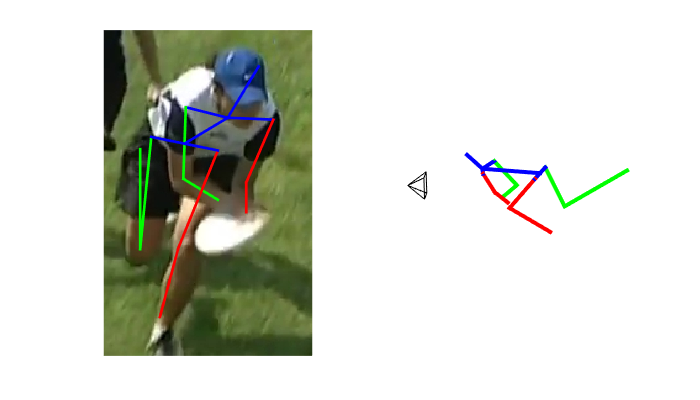}
	\includegraphics[width=0.32\linewidth,trim={2.5cm .7cm 0 .7cm},clip]{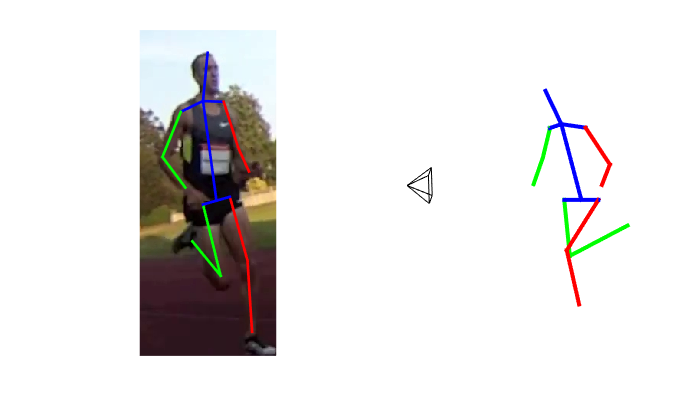}\\
	\includegraphics[width=0.32\linewidth,trim={2.5cm .7cm 0 .7cm},clip]{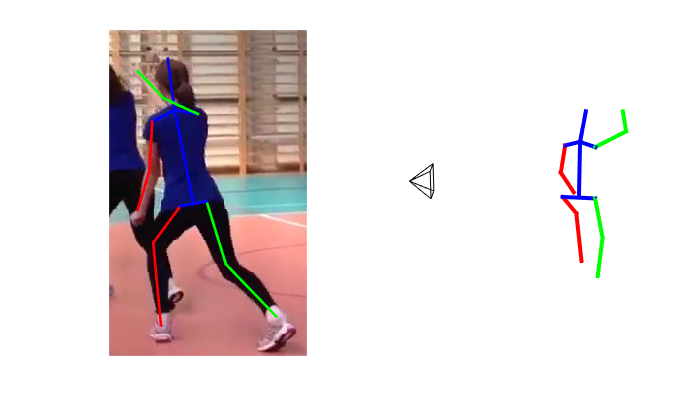}
	\includegraphics[width=0.32\linewidth,trim={2.5cm .7cm 0 .7cm},clip]{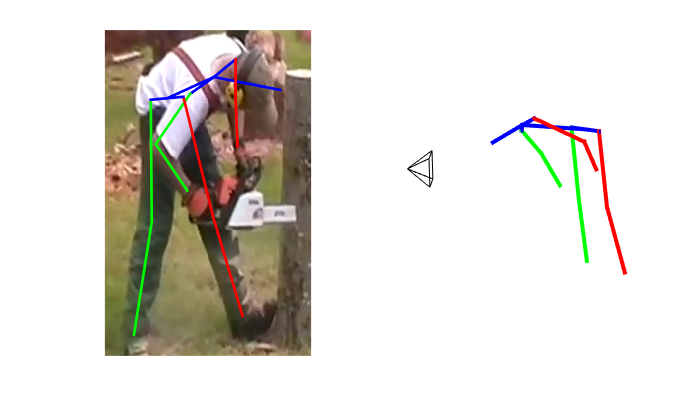}
	\includegraphics[width=0.32\linewidth,trim={2.5cm .7cm 0 .7cm},clip]{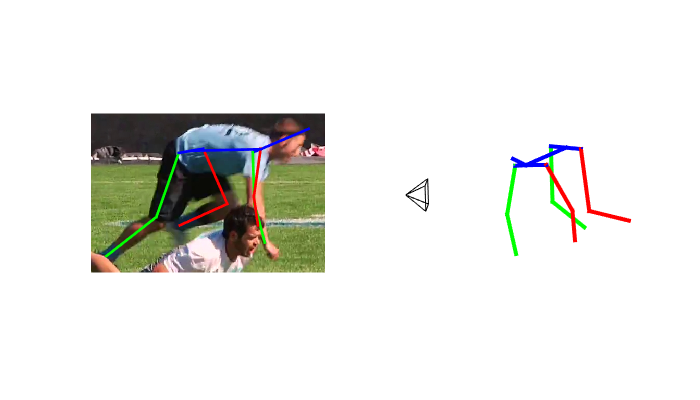}\\

	\caption{Examples of 3D pose reconstruction on images from MPII dataset (no ground-truth 3D skeleton). Each image shows overlaid 2D pose and the estimated 3D skeleton.}
	\vspace{-1ex}
	\label{fig:mpii}
\end{figure}

\begin{figure}[htb!]
	\centering
	\includegraphics[width=0.32\linewidth,trim={2.5cm .7cm 0 .7cm},clip]{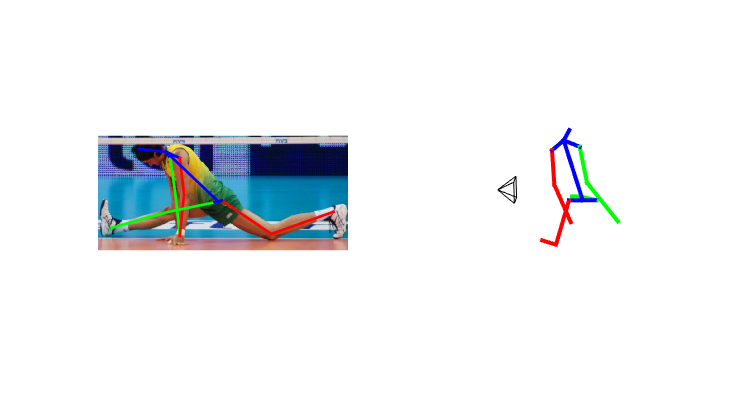}
	\includegraphics[width=0.32\linewidth,trim={2.5cm .7cm 0 .7cm},clip]{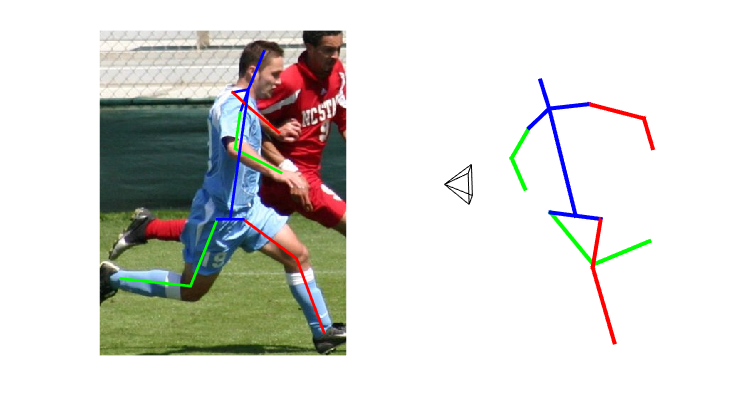}	
	\includegraphics[width=0.32\linewidth,trim={2.5cm .7cm 0 .7cm},clip]{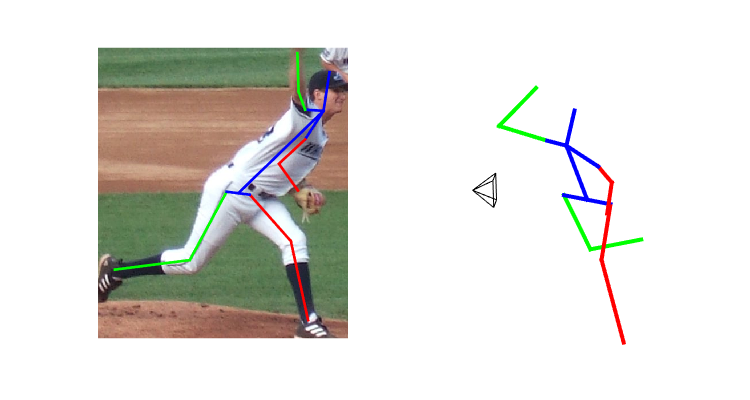}\\
	\includegraphics[width=0.32\linewidth,trim={2.5cm .7cm 0 .7cm},clip]{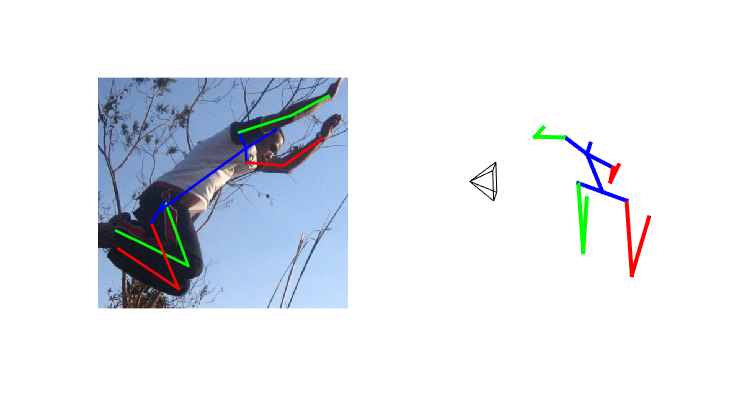}
	\includegraphics[width=0.32\linewidth,trim={2.5cm .7cm 0 .7cm},clip]{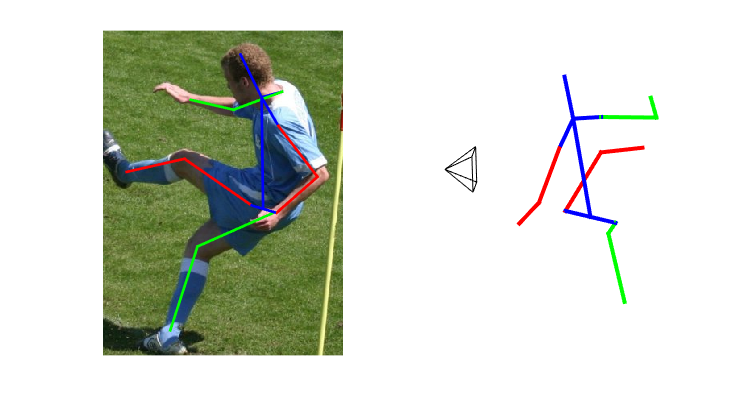}
	\includegraphics[width=0.32\linewidth,trim={2.5cm .7cm 0 .7cm},clip]{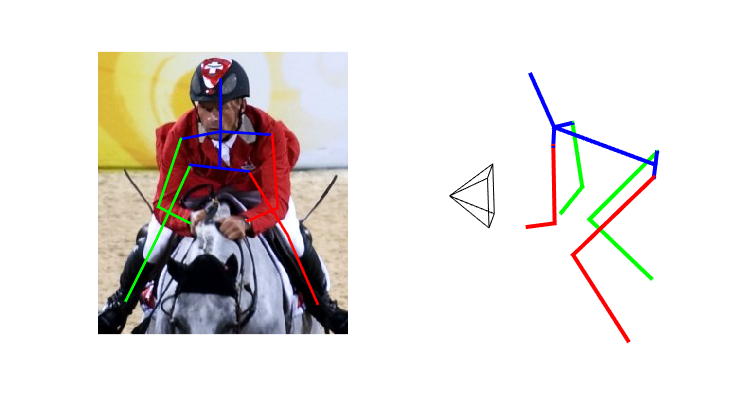}
	
	\caption{Examples of 3D pose reconstruction on images from LSP dataset (no ground-truth 3D).}
	\label{fig:leeds}
\vspace{-3ex}
\end{figure}

\subsection{Qualitative Results}
Figure~\ref{fig:h36good} shows some of the 3D pose reconstruction results on Human3.6M dataset using our lifting network. The ground truth 3D skeleton is depicted in gray. 
Some of the failures are shown in Figure~\ref{fig:h3failure}. Most of these can be attributed to self-occlusions or flip ambiguities in viewing direction (for more details see Suppl. materials).

To demonstrate generalization, we show some examples of 3D skeletons estimated on MPII~\cite{andriluka14cvpr} and the Leeds Sports Pose (LSP)~\cite{johnson2010clustered} datasets, in Figures~\ref{fig:mpii} and~\ref{fig:leeds} respectively. MPII has images extracted from short Youtube videos. LSP dataset consists of images of sport activities sampled from Flickr. Our unsupervised method successfully recovers 3D poses on these datasets without being trained on them. 

%% file: ResultTable_Abalation.tex
\begin{table}[htb!]
\centering
\begin{tabularx}{\linewidth}{lXrr}
	\toprule
	Supervision & Algorithm & \multicolumn{2}{c}{Error (mm)} \\
	 &  & GT & IMG\\
	\midrule \midrule
	Full & Chen~\etal~\cite{ChenDeva2017}  & 57.5 & 82.7 \\
    & Martinez~\etal~\cite{MartinezICCV2017} & 37.1 & 52.1\\
	\midrule
	Weak & 3DInterpreter~\cite{InterpreterNetwork2016} & 88.6 & 98.4\\
	& AIGN~\cite{Tung_2017_ICCV}  & 79.0 & 97.2\\
	& Drover~\etal~\cite{ZedNet_2018_ECCVW} & 38.2 & 64.6 \\
	\midrule
	Unsupervised & Rhodin~\etal~\cite{Rhodin_2018_ECCV} & - & 98.2 \\
	&Ours & \textbf{51} & \textbf{68}\\
	\bottomrule
\end{tabularx}
\caption{Comparison to the state-of-the-art unsupervised method of Rhodin~\etal~\cite{Rhodin_2018_ECCV} on Human3.6M. Comparable metrics for fully/weakly supervised methods are included for reference. Our approach outperforms~\cite{Rhodin_2018_ECCV} and several weakly supervised approaches~\cite{Tung_2017_ICCV,InterpreterNetwork2016} by a significant margin. GT and IMG denote results using ground truth 2D pose and estimated 2D pose by SH/CPM~\cite{stacked-hourglass,cpm}, respectively.}
\label{table:result_summary}
\end{table}

\begin{table}[htb!]
	\centering
	\begin{tabularx}{\linewidth}{lXrrr}
		\toprule
		Supervision & Algorithm &Trainset& PCK & AUC \\
		\midrule \midrule
		Full & Mehta~\cite{mono-3dhp2017} & MPI& 72.5 & 36.9\\
		& Mehta~\cite{mono-3dhp2017} & H36M& 64.7 & 31.7\\ 
       \midrule
       Weak & Zhou~\cite{zhou2017towards} &H36M & 69.2 & 32.5 \\
       \midrule
       Unsupervised & Ours & MPI& \textbf{71.1} & \textbf{36.3}\\
       & Ours & H36M& \textbf{64.3} & \textbf{31.6}\\
       \bottomrule
    \end{tabularx}
\caption{Our results (14-joint) on MPI-INF-3DHP with metrics as in Mehta~\etal~\cite{mono-3dhp2017}. The proposed unsupervised approach achieves similar performance as ~\cite{mono-3dhp2017} and ~\cite{zhou2017towards}.}
\label{table:MPI}
\end{table}   

\begin{table}[htb!]
\centering
\begin{tabularx}{\linewidth}{lXc}
	\toprule
	Type & Ablation & Error (mm) \\
	\midrule \midrule
	Architecture/ &SS & 162 \\
Loss Variations	&SS + Symm  & 168 \\
	&Adv& 61 \\
	&Adv+DA & 59 \\
	&Adv+SS & 58 \\
	&Adv+SS+DA & 55 \\
	&Adv+SS+DA+TD & \textbf{51} \\
	\midrule
	Supervised Fine-Tuning &0\% & 55 \\
with 3D Data	&5\%  & 37 \\
	\bottomrule
\end{tabularx}
\caption{Ablation studies.The architecture/loss ablations show the effect of various components on the unsupervised training. Supervised fine-tuning with only 5\% of randomly sampled Human3.6M 3D data gives similar performance as compared to fully-supervised results of~\cite{MartinezICCV2017}.}
\label{table:result_ablation}
\vspace{-3ex}
\end{table}

%% file: conclusions.tex
\subsection{Discussion}
Previous unsupervised and weakly supervised methods use additional constraints on training data in lieu of 3D annotations. For example,~\cite{ZedNet_2018_ECCVW, Yasin_2016_CVPR} leverage synthetic 2D poses obtained from known 3D skeletons to improve results. 
Similarly, Rhodin~\etal~\cite{Rhodin_2018_ECCV} derive an appearance and geometric model by choosing different frames from temporal sequences and multi-view images involving the same person. However, in theory, if multi-view images from synchronized cameras are available, one could triangulate the detected 2D joints to get 3D joints and train a supervised network. 
In contrast, our method treats each 2D skeleton as an individual training example, without requiring any multi-view correspondence. Hence, there is no restriction on where the 2D input pose originates; it could be obtained from a single image, video, or multi-view sequences. Our work explores the innate geometry of human pose itself, whereas~\cite{Rhodin_2018_ECCV} exploits the consistency in camera geometry and appearance of specific individuals. As shown in Sect.~\ref{subsect:quant_results}, our approach is able to augment the training data from other datasets (\eg Kinetics) with 2D skeletons captured in the wild.

Our current approach cannot handle occluded/missing joints during training or testing phases. This limits the amount of external domain data that can be used for training. For example, using OpenPose on Kinetics dataset results in 17M skeletons with at least 10 joints, but only 9M complete skeletons (14 joints). Though not the main focus of the paper, we did a small experiment to fill-in missing joints to further augment our training data. We trained a two-layer fully connected neural network which takes incomplete OpenPose 2D pose estimates on Human3.6M images as input and outputs completed 14 joints. The network was trained using the corresponding 2D ground-truth joints from Human3.6M in a supervised manner. Using the completed poses (17M skeletons) from the Kinetic dataset, our method achieved a MPJPE of 48mm on Human3.6M test data. This experiment further underscores the importance of volume and diversity of training data for unsupervised learning. We believe that by using auto-encoders and other unsupervised methods for data completion will enable utilizing even more diverse datasets, where 2D joints may be extracted from a variety of 2D pose estimation algorithms. Future work includes training the filling network and the domain adaptation network together with the lifting network in an end-to-end manner.

\section{Conclusions}
\label{sect:conclusions}

For 3D human pose estimation, acquiring 3D MoCap data remains an expensive and challenging endeavor. We presented an unsupervised learning approach to generate 3D skeletons from 2D joints, which does not require 3D data in any form. Our paper introduces geometric self-supervision as a novel constraint to learn the 2D-3D lifter. We showed that while geometric self-supervision is not a sufficient condition and cannot generate realistic skeletons by itself, it improves the reconstruction accuracy when combined with a discriminator. By training a domain adapter, we showed how to utilize data from different domains and datasets in an unsupervised manner. Thus, we believe that our paper has significantly improved the state-of-art in unsupervised learning of 3D skeletons by developing the key idea of geometric self-supervision and utilizing domain adaptation. Future work includes end-to-end training of 3D skeletons from 2D images, using self-supervision.